\documentclass[a4paper,fleqn]{cas-dc}

\usepackage[numbers]{natbib}
\usepackage{subfig}
\def\tsc#1{\csdef{#1}{\textsc{\lowercase{#1}}\xspace}}
\tsc{WGM}
\tsc{QE}

\begin{document}
\let\WriteBookmarks\relax
\def\floatpagepagefraction{1}
\def\textpagefraction{.001}

\shorttitle{Video Prediction by Efficient Transformers}    

\shortauthors{Ye, Bilodeau}  

\title [mode = title]{Video Prediction by Efficient Transformers}  



%

\author[1]{Xi Ye}[orcid=0000-0002-1763-6019]

\cormark[1]

\fnmark[]

\ead{xi.ye@polymtl.ca}

\ead[url]{}

\credit{}

\affiliation[1]{organization={LITIV Lab, Polytechnique Montréal},
            addressline={P.O. Box 6079, Station centre-ville}, 
            city={Montreal},
            postcode={H3C3A7}, 
            state={QC},
            country={Canada}}

\author[1]{Guillaume-Alexandre Bilodeau}[orcid=0000-0003-3227-5060]

\fnmark[]

\ead{gabilodeau@polymtl.ca}

\ead[url]{}

\credit{}

\cortext[1]{Corresponding author}



\begin{abstract}
Video prediction is a challenging computer vision task that has a wide range of applications. In this work, we present a new family of Transformer-based models for video prediction. Firstly, an efficient local spatial-temporal separation attention mechanism is proposed to reduce the complexity of standard Transformers. Then, a full autoregressive model, a partial autoregressive model and a non-autoregressive model are developed based on the new efficient Transformer. The partial autoregressive model has a similar performance with the full autoregressive model but a faster inference speed. The non-autoregressive model not only achieves a faster inference speed but also mitigates the quality degradation problem of the autoregressive counterparts, but it requires additional parameters and loss function for learning. Given the same attention mechanism, we conducted a comprehensive study to compare the proposed three video prediction variants. Experiments show that the proposed video prediction models are competitive with more complex state-of-the-art convolutional-LSTM based models.
The source code is available at \emph{https://github.com/XiYe20/VPTR}.
\end{abstract}




\maketitle

\section{Introduction}

Video future frames prediction (VFFP) can be applied to many application areas, for example, autonomous vehicles to predict future traffic outcomes \cite{bolte2019}, anomaly detection ~\cite{liu2018a} and model-based reinforcement learning \cite{leibfried2016} to predict users motion. Besides, VFFP is naturally a good self-supervised learning task, and it has drawn much attention in recent years \cite{bengio2013, wang2015}. Indeed, we can train VFFP models to learn good spatio-temporal representations from a large amount of unlabeled videos, and then apply it for many downstream tasks.

In this paper, we focus on the most common video prediction task, i.e. predicting $N$ future frames given $L$ past frames, with $L$ and $N$ being greater than 1. From the perspective of learning a deep neural networks for VFFP, we can formalize the task to be
\begin{equation}
    \arg\max_\theta p({\hat{x}_{L+N}, ..., \hat{x}_{L+1}}|x_L, ..., x_1; \theta),
\end{equation}
where $x_t$ and $\hat{x}_t$ denote the input past frames and predicted future frames respectively, $\theta$ denotes the parameters of the neural networks.

Benefiting from the advances in deep learning, the performance of VFFP models is constantly improving. Particularly, the efficient and powerful Convolutional Long Short-Term Memory networks (ConvLSTMs) are the core for almost all the state-of-the-art (SOTA) VFFP models. Nevertheless, ConvLSTMs suffer from some inherent problems typical of recurrent neural networks (RNNs), including slow training and inference speed, error accumulation during inference, vanishing gradient, and predicted frame quality degradation. Researchers keep improving the performance by developing more and more sophisticated ConvLSTM-based models. For instance, by integrating custom motion-aware units into ConvLSTM \cite{chang2021}, or building complex memory modules to store the motion context \cite{lee2021}. Still, VFFP is a challenging task which is far from being solved. 

Meanwhile, the Transformer \cite{vaswani2017} overcomes most of the aforementioned drawbacks of RNNs and made breakthroughs for natural language processing (NLP). Inspired by this, more and more researchers are starting to adapt the Transformer for various computer vision tasks \cite{dosovitskiy2021, esser2021, Arnab_2021_ICCV}, including few recent works for VFFP \cite{liu2020e,yan2021a, wu2021b}. However, the Transformer was originally designed for sequence data processing. Given a sequence of elements with length $L$ and feature dimensionality $d_{model}$, a dot product attention-based Transformer has a complexity of $\mathcal{O}(L^2d_{model})$. So it is computationally expensive to apply a Transformer to high dimensional visual features. For example, considering a spatiotemporal feature $Z \in R^{T\times H \times W \times d_{model}}$, where $T, H, W$ denote the time duration, spatial height and spatial width respectively, we need to flatten it to be a sequence $Z \in R^{(T\cdot H\cdot W) \times d_{model}}$ with length $T\cdot H\cdot W$ along the spatial and temporal dimensions for a standard Transformer. The complexity is therefore $\mathcal{O}((THW)^2d_{model})$. We still need further research about more efficient and compact visual Transformer, especially for videos. \emph{Therefore, we propose a novel efficient Transformer block with smaller complexity, named VidHRFormer, and we developed a new video prediction Transformer (VPTR) based on it.}

Among the Transformer-based VFFP models \cite{liu2020e,yan2021a, wu2021b} that we mentioned earlier, some of them are autoregressive models while some others are non-autoregressive models, and they are based on different attention mechanisms, e.g. a custom convolution multi-head attention (MHA) \cite{liu2020e} and standard dot-product MHA \cite{yan2021a, wu2021b}. There is no formal comparison of the two typical approaches (autoregressive vs non-autoregressive) in the case of Transformer-based VFFP models so far. \emph{Thus, we developed a non-autoregressive VPTR model (VPTR-NAR), two autoregressive VPTR variants, i.e., the fully autoregressive VPTR model (VPTR-PAR) and the partial autoregressive VPTR model (VPTR-FAR).} All VPTR variants share the same attention mechanism and same number of Transformer block layers, which guarantees a fair comparison between the two approaches.

Our main contributions are summarized as:

1) We proposed a new efficient Transformer block, VidHRFormer, for spatio-temporal feature learning by combining spatial local attention and temporal attention in two steps. The new Transformer block successfully reduces the complexity of a standard Transformer block with respect to the same input spatio-temporal feature size, specifically, from $\mathcal{O}((THW)^2d_{model})$ to $\mathcal{O}((\frac{H^2W^2}{P^2} + T^2)d_{model})$.

2) Based on this Transformer block, three VPTR models, VPTR-NAR, VPTR-FAR and VPTR-PAR, were developed. We show that the proposed simple attention-based VPTRs are capable of reaching and outperforming more complex SOTA ConvLSTM-based VFFP models.

3) A formal comparison of three VPTR variants was conducted. The results show that VPTR-NAR has a faster inference speed and smaller accumulation of errors during inference, but it is more difficult to train. We solved the training problem of VPTR-NAR by employing a contrastive feature loss that maximizes the mutual information of predicted and ground-truth future frame features.

4) We found that given the same number of Transformer block layers, VPTR-FAR and VPTR-PAR have a worse generalization performance due to the accumulated inference errors, which are introduced by the discrepancy between train and test behaviors. Finally, we compared two different inference methods for VPTR-FAR and VPTR-PAR, the results show that recurrent inference over pixel space introduces less accumulation of errors than recurrent inference over latent space in the case of VPTR-FAR.

This paper is an extension of the work in \cite{ye2022}. We present a new VPTR-PAR variant and we fully analyze auto-regressive vs non-auto-regressive models for our new VidHRFormer.

\section{Related work}
\subsection{Video future frames prediction}
Almost all the deep learning-based VFFP models take an encoder to extract the representations of past frames, and then a decoder generates future frame pixels based on those representations. Here, we propose our own taxonomy for VFFP models, which is derived from the different encoding and decoding mechanisms. 

\textbf{Non-decomposing and Decomposing models}. Most VFFP models are equipped with an encoder which generates frame-level features. they are classified as non-decomposing models since they assume that the encoders implicitly extract all the necessary information for the prediction of future frames \cite{jin2020, wang2020, jayaraman2019, babaeizadeh2018, denton2018}. However, some other VFFP models explicitly decompose the frame visual features, e.g. content and motion, which aims to reduce the difficulty of prediction by introducing more prior knowledge \cite{villegas2017a, franceschi2020, jang2018, tulyakov2018, denton2017}. There are other decomposing methods, including object keypoint skeleton and appearance decomposition \cite{cai2018, fushishita2020}, object-centric decomposition \cite{hsieh2018, kosiorek2018a, chen2017}.

\textbf{Sequentially-generated and Parallelly-generated models}. The majority of VFFP models rely on the flexible RNNs for temporal information modeling, which predict future frames recursively. We call these models sequentially-generated models \cite{jayaraman2019, babaeizadeh2018, castrejon2019, denton2018, kwon2019}.
In contrast, some models take advantage of standard CNNs or 3D-CNNs as the backbones, without the use of RNNs, and generate multiple future frames simultaneously (that is in parallel) \cite{mathieu2016, chen2017a, vondrick2016, wu2020a}. 

\textbf{Pixel-direct generation and Transformation-based models}. Some VFFP models use decoders to directly synthesize the pixels of future frames based on past frame representations \cite{chen2020, franceschi2020, kwon2019, cai2018, villegas2017a}, while some other models have no such decoder and generate the future frames by applying spatial transformation operation on the past frames, like warping \cite{chen2017, jin2018, villegas2017a}.

\textbf{Deterministic and Stochastic models}. The future prediction is by nature multimodal \cite{fragkiadaki2017}, i.e. stochastic, even though we are given multiple past frames as context. However, it is difficult to model the stochasticity of the future and thus most VFFP models ignore it, even though it affects the predicted image quality \cite{jin2020, wu2020a, chen2020}. Stochastic models normally make uncertainty estimation based on VAE or GAN, such as stochastic variational video prediction (SV2P) model \cite{babaeizadeh2018}, stochastic video generation with a learned prior (SVG-LP) \cite{denton2018}, and improved conditional variational RNNs (VRNNs) \cite{castrejon2019}.

Note that the categories described above are not mutually exclusive as a model can be a combination of arbitrary different encoders and decoders. For example, our proposed VPTR belongs to the deterministic, non-decomposing, pixel-direct generated model categories. The VPTR-NAR variant is considered to be a parallelly-generated model, while VPTR-PAR and VPTR-FAR variants are sequentially-generated models.

Recently, many work achieve SOTA performance by integrating attention mechanism or memory augmented modules in the ConvLSTM models for a better long-term motion information learning \cite{xu2020, lee2021, chang2021}. For example, LMC-Memory model \cite{lee2021} stores the long-term motion context by a novel memory alignment learning, and the motion information is recalled during test to facilitate the long-term prediction. Specifically, LMC-Memory takes the difference image between adjacent frames as motion information, same as MCnet \cite{villegas2017a}, to learn the long-term motion context embeddings as an external memory during training. During test, the recalled motion context memory feature is concatenated with spatial feature of current time step to predict the next frame. VPEG \cite{xu2020} shares a similar idea with LMC-Memory. VPEG also learns a pool of motion features, which are called examples, of the whole dataset, but VPEG aims to approximate the stochastic predictions by those examples and thus bypassing the variational inference of other aforementioned stochastic video prediction models. 

Zhang et al. \cite{chang2021} proposed an attention-based motion-aware unit (MAU) to increase the temporal receptive field of RNNs. The proposed MAU is composed of an attention module and a fusion module, where the attention module considers the correlation between the current step features and different history states within a temporal receptive field as an attention score, and the fusion module is responsible to aggregate those features. In this way, the MAU is able to capture better motion information and receive a broader temporal receptive field than the vanilla ConvLSTMs. However, those attention or memory augmented ConvLSTM architectures tend to be complex and it is hard to understand and follow the spatio-temporal information flow in the network. In this paper, we demonstrate that a highly modulated custom Transformer block is capable of capturing good spatio-temporal information for video prediction with a simpler network architecture.

\subsection{Transformers in computer vision}
The architecture of the proposed VPTR-NAR model follows the architecture of Detection Transformer (DETR) \cite{carion2020}, which relaxes the dependence on complex region proposal and non-maximal suppression in object detection by using Transformers. DETR follows the classical neural machine translation (NMT) Transformer architecture. The 2D image features extracted by a CNN are flattened into a sequence and fed into a standard Transformer encoder. The output features of the Transformer encoder, normally referred as memories, together with some learned object queries are fed into a Transformer decoder. The decoder output features, corresponding to each object query, are considered as the representation of each potential object, and finally a small feed forward neural network is used to do the classification or bounding box detection based on the extracted potential object features.

\textbf{Efficient visual Transformers.} The bottleneck of a basic Transformer comes from the quadratic complexity of the attention score calculation. There are mainly two types of models to reduce the computation cost for visual Transformers. The first class of models reduces the flattened sequence length by different methods. ViT and many successive works \cite{dosovitskiy2021, wang2021, Arnab_2021_ICCV} divide high resolution input into local patches, 2D or 3D, and then concatenate along the channel dimension to tokenize the patch. Pooling can be an alternative to sequence length reduction \cite{fan2021}. The second class of models introduces sparse attention to reduce the complexity, e.g. restricting the attention over a local region \cite{zhang2021, liu2021d, YuanFHLZCW21}, or decomposing the global attention into a series of axial-attention \cite{huang2019a, wang2020c, Arnab_2021_ICCV}. HRFormer \cite{YuanFHLZCW21} is an example of local region attention-based Transformer, which is designed for image classification and dense prediction. Essentially, HRFormer replaces the convolution layers of HRNet \cite{sun2019c} by self-attention. The multi-resolution parallel architecture is the same as HRNet. 

Our VidHRFormer block is inspired by the HRFormer block. Specifically, a HRFormer block is composed by a local-window multi-head self attention layer and a depth-wise convolution feed-forward network. The input feature maps $Z\in \mathbb{R}^{H \times W \times C}$ is firstly evenly divided into $P$ non-overlapping local patches, each patch is $Z_p\in \mathbb{R}^{\frac{H}{P} \times \frac{W}{P} \times C}$. Then a multi-head self-attention is performed for each patch. Finally, the depth-wise convolution is used to exchange information of different local patches. HRFormer is similar to the Swin Transformer \cite{liu2021d}. The Swin Transformer takes a cyclic shift window partitioning procedure instead of a depth-wise convolution layer to exchange the information of local patches.

\subsection{Transformers for VFFP}
A recent ConvTransformer \cite{liu2020e} model follows the architecture of DETR \cite{meinhardt2021}, which also inspired the development of our VPTR-NAR. Despite the similarities, our VPTR-NAR is different from ConvTransformer with respect to the fundamental attention mechanism. Specifically, ConvTransformer proposed a custom hybrid multi-head attention module that replaces both the linear projection and dot-product attention operation by a convolution, but our VPTR-NAR uses the standard multi-head attention module. The ConvTransformer attention map calculation has a complexity of $\mathcal{O}(HWTk_c^2d_{model}^2)$, where $k_c$ is the size of convolutional kernels. Given our experimental configurations, i.e., $H=8$ , $W=8$, $P=4$, $T=20$, and $d_{model}=512$, our efficient Transformer block (with a complexity of $\mathcal{O}((\frac{H^2W^2}{P^2} + T^2)d_{model})$) is more efficient. Another more recent VideoGPT \cite{yan2021a} model is capable of both single image animation and VFFP. VideoGPT takes a 3D CNN as backbone to encode video clips into spatial-temporal features, which are then flattened to be a sequence to train a standard Transformer with the GPT manner. VideoGPT shares a similar architecture and train/test behaviours as our VPTR-FAR, as both of them fully decompose a joint distribution into a product of conditional distribution, like GPT. But VideoGPT performs the attention along spatial and temporal jointly while our VPTR-FAR performs the attention along spatial and temporal separately with a smaller computational complexity. More importantly, VideoGPT takes a 3D CNN to downsample the time dimension of input videos to facilitate the temporal information modeling. In our case, we do not downsample. Our VPTR models solely depend on attention for a full temporal information modeling, without downsampling and loss of information. Besides, VideoGPT is a stochastic model based on VQ-VAE \cite{vandenoord2017}, instead of a deterministic model as our VPTR. Another recent work NÜWA \cite{wu2021b} shares a similar idea as VideoGPT.

\section{The proposed VPTR models}
\label{sec: methods}

\subsection{Overall framework of VPTR}
\begin{figure}[ht]
\centering
\includegraphics[clip, trim=5.9cm 8.8cm 5cm 15cm, width=\linewidth]{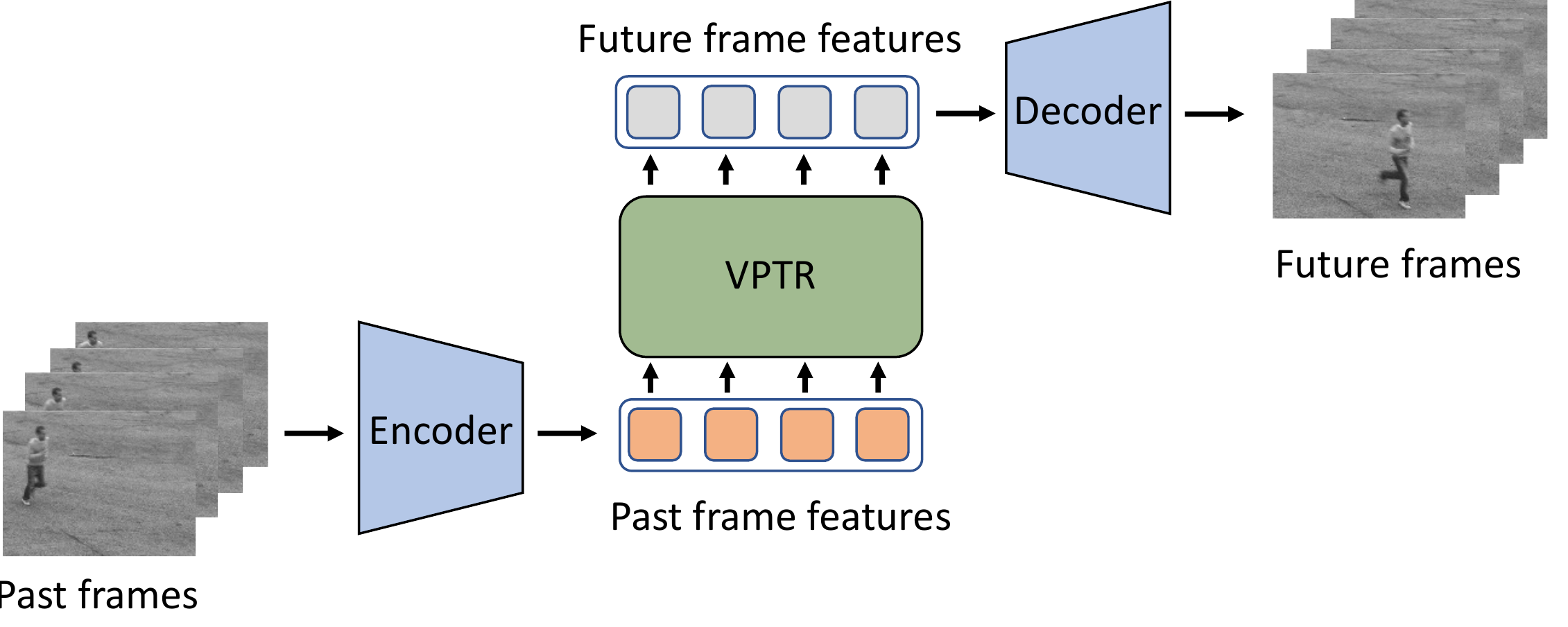}
\caption{Overall framework of the proposed VFFP model. VPTR predicts over latent feature space.}
\label{fig:VPTR overall framework}
\end{figure}

Our overall video prediction framework is illustrated in Fig. \ref{fig:VPTR overall framework}. It consists of an autoencoder and the VPTR module itself. Specifically, a CNN encoder shared by all the past frames extracts the visual features of each frame, then a VPTR, based on VidHRFormer blocks, is applied to predict the visual features of each future frame based on the past frame features. Finally, we reconstruct the pixels of each future frame with the CNN decoder. The detail architectures of the autoencoder and three different VPTR variants are described in the following subsections. 

In contrast to most ConvLSTM-based models, here we disentangle the visual feature extraction and the prediction process, similar to VideoGPT \cite{yan2021a} and VPEG \cite{xu2020}. The benefit of the disentanglement is that we are able to derive the explicit representations of the input past video clips, and use them for downstream tasks, like action recognition, early action prediction etc. In other words, because of the disentanglement, it is easier to extend VPTR to solve self-supervised learning tasks, in contrast to most ConvLSTM-based models, especially in the case of the Transformation-based models, due to the lack of explicit representations. We naturally introduce a two-stages training strategy because of the feature extraction and prediction disentanglement, which increases the flexibility and reduces the learning burden of VPTR. For simplicity, Fig. \ref{fig:VPTR overall framework} only shows the inference behavior, the two-stages training strategy is described in detail at the end of this section.

\subsection{Encoder and decoder}

We adapted the ResNet-based autoencoder from the Pix2Pix model \cite{Isola2017}. In order to match with the VPTR model feature dimensionality $d_{model}$, the only modification is that we change the encoder output feature channels and decoder input feature channels to be of size $d_{model}$. The reason to use Pix2Pix autoencoder is that there are no skip connections between the encoder and decoder. Normally, the encoder-decoder skip connections are good for a higher reconstruction image quality, like the famous U-Net architecture \cite{ronneberger2015}. However, it is incompatible with our strategy of disentangling feature extraction and prediction. We chose this strategy because we hypothesize that the learned visual features by the encoder will include all the information of the input and thus would be better for future self-supervised learning tasks, while skip connections would enable some information to bypass the bottleneck encodings. More importantly, we aim to predict future frames instead of simply reconstructing past frames. The encoder-decoder skip connections from past frame features to future frame features make the two-stages training impractical because future image transformations tend to be ignored and predictions are not learned. Training jointly does not solve this problem. This was confirmed by preliminary experiments. We failed to conduct a successful joint training of the Transformer and the autoencoder.

The loss function to train the encoder and decoder includes three terms and is given by,
\begin{multline}
    \mathcal{L}_{rec} = \mathcal{L}_2(X,\hat{X}) + \mathcal{L}_{gdl}(X, \hat{X})\\
    + \lambda_1 \arg\min_G\max_D  \mathcal{L}_{GAN}(G, D),
\label{eq: AE_loss}
\end{multline}
where $\mathcal{L}_2$ denotes the MSE loss (Eq. \ref{eq:l2}), $\mathcal{L}_{gdl}$ denotes image gradient difference loss \cite{mathieu2016} (Eq. \ref{eq:gdl}) and $\mathcal{L}_{GAN}$ denotes the GAN loss (Eq. \ref{eq:gan_loss}). Both $\mathcal{L}_{gdl}$ and $\mathcal{L}_{GAN}$ are utilized to learn the details in the images and thus provide a higher visual quality. $X$ and $\hat{X}$ denote the original frames and reconstructed frames respectively, $x_i$ denotes a single frame, $\lambda_1$ and $\alpha$ are hyperparameters. In Eq. \ref{eq:gan_loss}, $D$ denotes a discriminator, which is not shown in Fig. \ref{fig:VPTR overall framework}, and the combination of the encoder and decoder is considered to be a generator $G$. We train $\mathcal{L}_{GAN}$ with the PatchGAN \cite{Isola2017} manner. 

\begin{equation}
    \mathcal{L}_2(X,\hat{X}) = \sum_{i=1}^n \lVert x_i-\hat{x}_i \rVert _2^2
\label{eq:l2}
\end{equation}

\begin{multline}
    \mathcal{L}_{gdl}(X,\hat{X}) =\sum_{i=1}^n \sum_{i,j} \big | \lvert {x_{i,j}}-x_{i-1, j} \rvert - \lvert {\hat{x}_{i,j}}-\hat{x}_{i-1, j} \rvert \big |^\alpha\\
    + \big | \lvert {x_{i,j-1}}-x_{i, j} \rvert - \lvert {\hat{x}_{i,j-1}}-\hat{x}_{i, j} \rvert \big |^\alpha
\label{eq:gdl}
\end{multline}

\begin{equation}
    \mathcal{L}_{GAN}(G, D) = \mathbb{E}_X[logD(X)] + \mathbb{E}_{\hat{X}}[log(1 - D(G(X))]
\label{eq:gan_loss}
\end{equation}

\subsection{VidHRFormer Block}

\begin{figure*}[ht]
\centering
\subfloat[][]{
\includegraphics[clip, trim=11cm 7.35cm 11.5cm 7cm, width=0.183\linewidth]{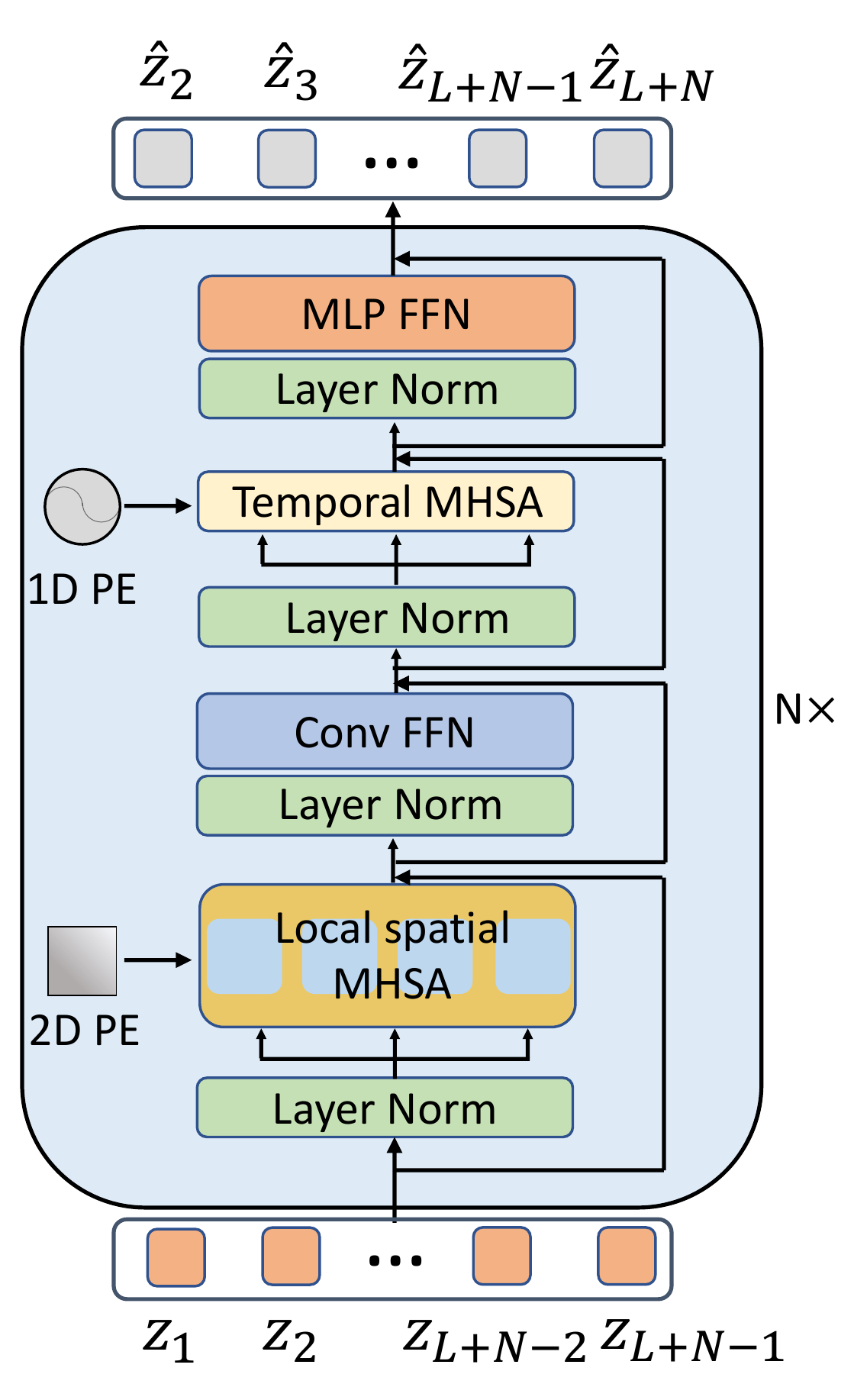}
\label{fig:FAR_model}}
\subfloat[][]{
\includegraphics[clip, trim=3.5cm 3cm 7.7cm 5.2cm, width=0.366\linewidth]{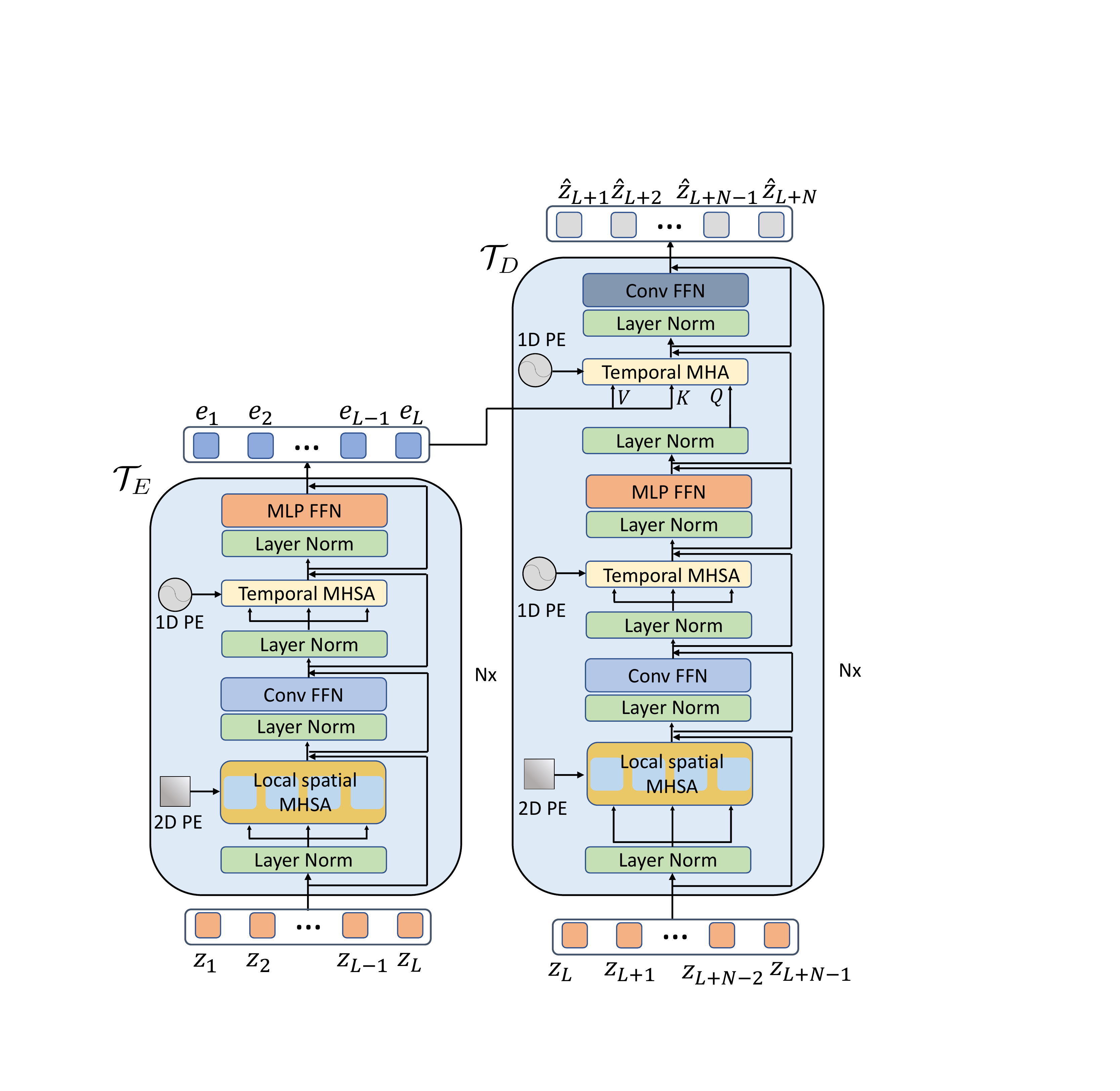}
\label{fig:PAR_model}}
\subfloat[][]{
\includegraphics[clip, trim=3.5cm 3cm 7.7cm 5.2cm, width=0.366\linewidth]{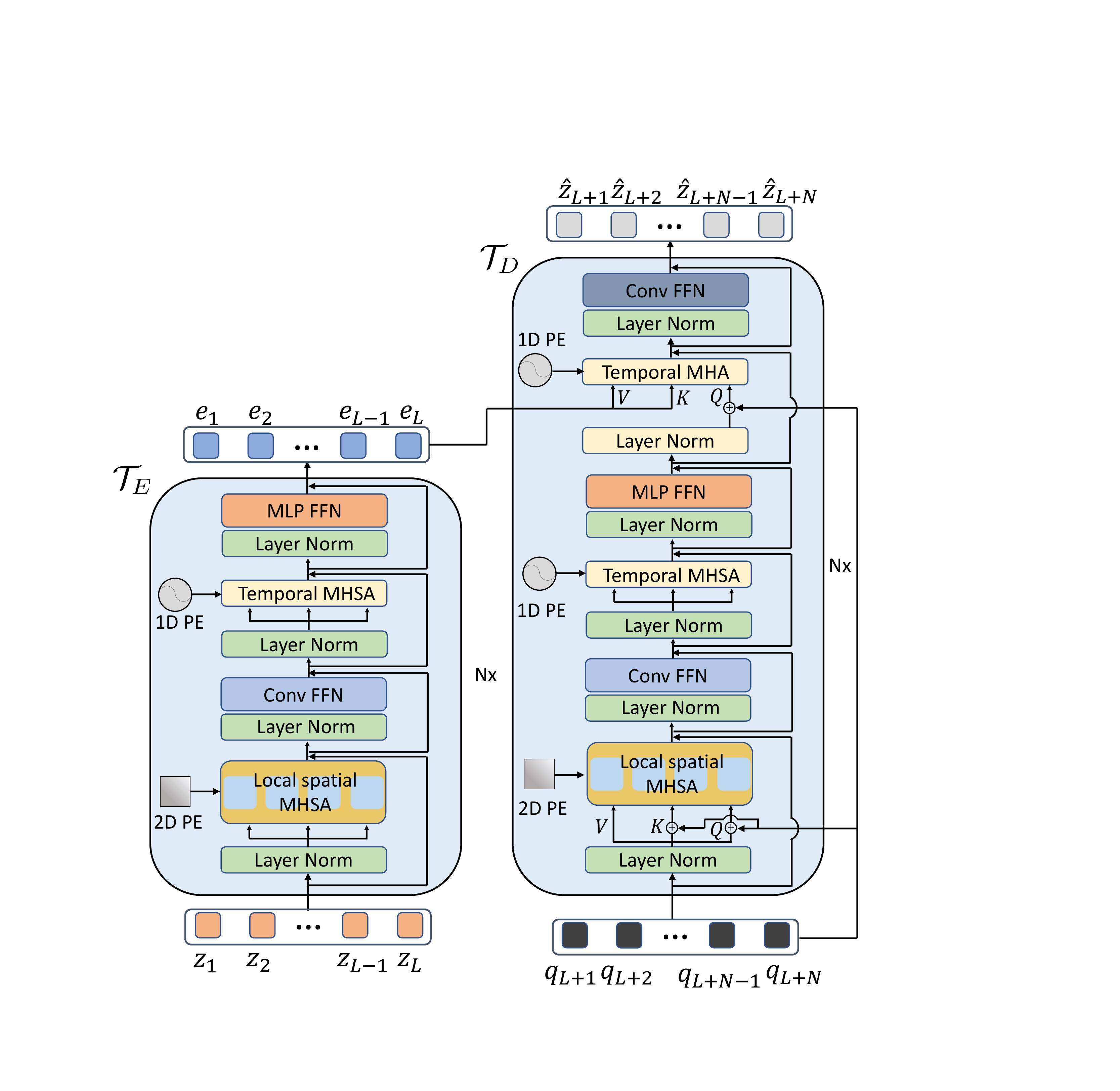}
\label{fig:NAR_model}}

\caption{(a) VPTR-FAR. The blue area indicates the proposed basic VidHRFormer block. A temporal attention mask is applied to the Temporal MHSA module for VPTR-FAR. (b) VPTR-PAR. There is an attention mask for the Temporal MHA layer of the autoregressive Transformer decoder $\mathcal{T}_D$. (c) VPTR-NAR. The  Transformer decoder $\mathcal{T}_D$ is non-autoregressive.}
\end{figure*}

In order to reduce the complexity of a standard Transformer and to make it practical for high-dimensional video representation learning, our solution is to apply attention only over local spatial patch and separate the spatial and temporal attention. The proposed new video representation learning Transformer block is named as VidHRFormer block, see the blue area of Fig. \ref{fig:FAR_model} for the detail architecture. Essentially, we extend the  HRFormer block \cite{YuanFHLZCW21} by integrating a temporal multi-head attention layer, together with some other necessary feed-forward and normalization layers.

\textbf{Local spatial multi-head self-attention (MHSA).} Considering a batch of video feature maps $Z \in \mathbb{R}^{N\times T\times H \times W \times d_{model}}$, the local spatial MHSA is shared by frames at different time steps. So $Z$ is firstly reshaped and evenly divided into $P= \frac{HW}{K^2}$ local patches $\{Z_1, Z_2, ..., Z_P\}$ along the $H$ and $W$ dimensions, where each local patch is of size $K\times K$, therefore $Z_p \in \mathbb{R}^{(NT) \times K^2 \times d_{model}}$. The multi-head self-attention is formulated as $MHSA(Z_p) = Concat[head(Z_p)_1, ..., head(Z_p)_h]$, where $Concat$ denotes the concatenation operation and $head(Z_p)_i \in \mathbb{R}^{K^2\times \frac{d_{model}}{h}}$ is calculated by

\begin{equation}
    head(Z_p)_i = softmax[\frac{((Z_p^Q W_i^Q)(Z_p^K W_i^K)}{\sqrt{d_{model}/h}}]Z_p W_i^V,
\label{eq: attention}
\end{equation}

\noindent where $W_i^Q$, $W_i^K$, $W_i^V$ are linear projection matrices for the query, key and value  of each head $i$ respectively, $Z_p^Q$ and $Z_p^K$ denote the key and query for attention.  The complexity of local spatial MHSA is $\mathcal{O}(\frac{H^2W^2}{P^2}d_{model})$.

Another critical component for a Transformer is the positional encoding, which enables the Transformer to output different representation of the same input element given different context. Both fixed absolute 2D positional encoding \cite{carion2020} and relative positional encoding (RPE) \cite{shaw2018} are good candidates for the local patch to get $Z_p^Q$ and $Z_p^K$. These two different positional encodings are compared in the experiments. 

\textbf{Convolutional feed-forward neural network (Conv FFN).} Conv FFN is also shared by frames at different time step of the video. The outputs of the local spatial MHSA, $\{Z_1, Z_2, ..., Z_P\}$ are assembled back to be $Z \in \mathbb{R}^{(NT)\times H \times W \times d_{model}}$. The Conv FFN layer includes a $3\times 3$ depth-wise convolution and two point-wise MLPs. Different from the the original HRFormer block, all the normalization layers in Conv FFN are layer normalization, instead of batch normalization.

\textbf{Temporal MHSA.} A temporal MHSA is placed on top of the previously described local spatial MHSA and Conv FFN to model the temporal dependency between frames. It is shared by features across different spatial locations. In other words, the input feature map $Z \in \mathbb{R}^{(NT)\times H \times W \times d_{model}}$ is reshaped to be $Z \in \mathbb{R}^{(NHW)\times T \times d_{model}}$. The temporal MHSA uses the same standard multi-head self-attention as the local spatial MHSA and thus the complexity of temporal MHSA is $\mathcal{O}(T^2d_{model})$. However, here, for simplicity, we selected a fixed absolute 1D positional encoding for the time steps. 
Same as the standard Transformer, we place a MLP feed-forward neural network after the temporal MHSA. Finally, the output feature map is reshaped back to be $Z \in \mathbb{R}^{N\times T\times H \times W \times d_{model}}$ to be used by the next VidHRFormer layer.

\begin{figure}[ht]
\centering
\includegraphics[width=\linewidth]{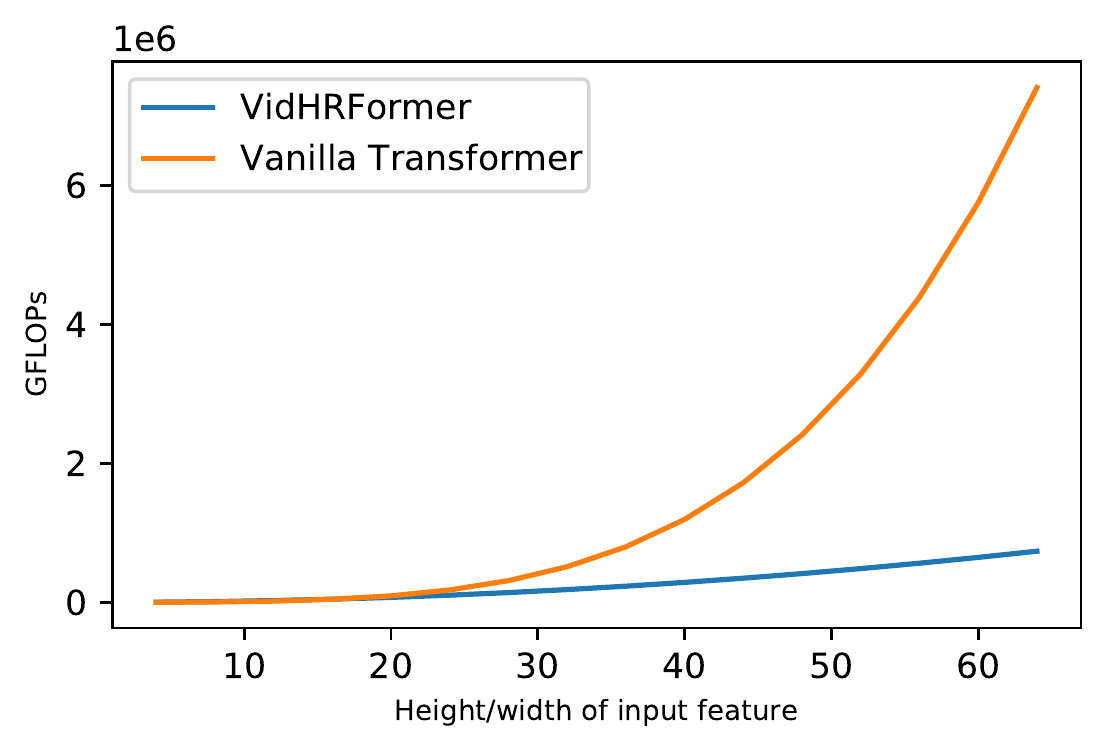}
\caption{GFLOPs comparison of a VidHRFormer block and a vanilla Transformer block.}
\label{fig:GFLOPs_comparison}
\end{figure}

To sum up, the total complexity of VidHRFormer block is the combination of spatial local window MHSA complexity and temporal MHSA complexity, i.e., $\mathcal{O}((\frac{H^2W^2}{P^2} + T^2)d_{model})$. This results in a better efficiency as compared to a standard Transformer that as a complexity of $\mathcal{O}((THW)^2d_{model})$. For a better demonstration that the proposed VidHRFormer block is much more efficient than a vanilla Transformer block, we plotted the GFLOPs curves w.r.t an increasing $H$ or $W$ ($T$ is fixed) in Figure \ref{fig:GFLOPs_comparison}. As the height or width of the input feature increases, the GFLOPs of a vanilla Transformer quickly increase and makes it infeasible for application. We would observe a similar phenomenon if we fix the $H$/$W$ and vary the video feature length $T$.

In the following sections, we describe three different VPTR models that are based on the VidHRFormer block.

\subsection{VPTR-FAR (fully autoregressive)}
\label{ssec: VPTR-FAR}
Together with a well-trained CNN autoencoder, the VPTR-FAR parameterizes the following distribution:
\begin{equation}
    p(x_1, ...,x_{L}, ..., x_{L+N}) = \prod_{t=1}^{L+N} p(x_t|x_{t-1}, ... x_1)
\end{equation}

In other words, VPTR-FAR predicts the next frame conditioned on all previous frames. This is the most common paradigm for most SOTA VFFP models. To enforce the causal relationship between the next frame and previous frames, an attention mask is applied to the temporal MHSA module. The fully autoregressive VPTR model is simply composed of a stack of multiple VidHRFormer blocks, see Fig. \ref{fig:FAR_model}.

For training, we feed the ground-truth frame feature sequence $\{z_1, ..., z_{L+N-1}\}$ generated by the encoder into VPTR-FAR, which then predicts the future feature sequence $\{\hat{z}_2, ..., \hat{z}_{L+N}\}$ for the decoder to generate frames $\{\hat{x}_2, ..., \hat{x}_{L+N}\}$. We define the training loss of VPTR-FAR as follows (see Eq. \ref{eq:l2} and \ref{eq:gdl}):
\begin{equation}
    \mathcal{L}_{FAR} = \sum_{t=2}^{L+N}\mathcal{L}_2(x_t,\hat{x_t}) + \sum_{t=2}^{L+N} \mathcal{L}_{gdl}(x_t, \hat{x_t})
\label{eq: FAR_loss}
\end{equation}

For test, given the ground-truth feature sequence $\{z_1, ..., z_{L}\}$ of all past frames, there are two different options for the recurrent prediction of the future frames. The first one is only taking the VPTR module to recurrently predicting all the future frame features, i.e. $\hat{z}_{t} = \mathcal{T}(z_1, ..., z_{t-1}), t\in[L+1, ... L+N]$, where $\mathcal{T}$ denotes the VPTR-FAR module. Then we get $\hat{x}_t = Dec(\hat{z}_t), t\in[L+1, ... L+N]$, where $Dec$ denotes the CNN frame decoder. The second option includes two additional steps. We firstly decode one future feature to be frame $\hat{x}_t$, and then encode the frame back into a latent feature before the prediction of next future frame feature, i.e., $\hat{z}_{t} = Enc(Dec(\mathcal{T}(z_1, ..., z_{t-1}))), t\in[L+1, ... L+N]$, where $Enc$ denotes the CNN frame encoder. The second method significantly reduces the accumulated test error during inference, and we analyze the reasons in the experiments section.
 
 \subsection{VPTR-PAR (partially autoregressive)}
The partially autoregressive variant is illustrated in Fig. \ref{fig:PAR_model}. It consists a Transformer encoder and decoder, where the encoder $\mathcal{T}_E$ encodes all past frame features $z_t, t\in[1, L]$ to be memories $e_{t}, t\in[1, L]$. The architecture of $\mathcal{T}_E$, left part of Fig. \ref{fig:PAR_model}, is the same as the VPTR-FAR, except that there is no temporal attention mask for the temporal MHSA module. The autoregressive Transformer decoder $\mathcal{T}_D$ of VPTR-PAR, right part of Fig. \ref{fig:PAR_model}, includes two more layers compared with $\mathcal{T}_E$, a temporal multi-head attention (MHA) layer and another output Conv FFN layer. The Temporal MHA layer is also called the encoder-decoder attention layer, which takes the memories as value and key, while the query is derived from the $\{z_L, ... z_{L+N-1}\}$. A temporal attention mask is applied to Temporal MHSA layer of $\mathcal{T}_D$ to achieve the autoregressive modeling. Theoretically, VPTR-PAR models the following distribution:
\begin{multline}
    p(x_{L+N}, ..., x_{L+1}|x_{L}, ...,x_1) = \\
    \prod_{t=L+1}^{L+N} p(x_t|x_{t-1}, ... x_{L+1}, x_L, ..., x_1)
\end{multline}

Compared with VPTR-FAR, we only decompose the probability distribution of future frames to be the product of a series of conditional distributions, so this model is named as partially autoregressive VPTR. During training, the frame features $\{z_1, ... z_{L+N-1}\}$ are divided into two parts, as shown in Fig. \ref{fig:PAR_model}, and fed into $\mathcal{T}_E$ and $\mathcal{T}_D$ respectively. Following the convention of NMT research, the input of $\mathcal{T}_E$ is called source sequence, and input of $\mathcal{T}_D$ is called target sequence. The training loss of VPTR-PAR is formulated as  

\begin{equation}
    \mathcal{L}_{PAR} = \sum_{t=L+1}^{L+N}\mathcal{L}_2(x_t,\hat{x_t}) + \sum_{t=L+1}^{L+N} \mathcal{L}_{gdl}(x_t, \hat{x_t}),
\label{eq: PAR_loss}
\end{equation}
where $\mathcal{L}_{gdl}$ and $\mathcal{L}_2$ are defined in Eq. \ref{eq:gdl} and Eq. \ref{eq:l2} respectively.

Same as VPTR-FAR, we can use the same two different recurrent inference methods for VPTR-PAR.

\subsection{VPTR-NAR (non-autoregressive)}
In order to reduce the predicted frames accumulated error and increase the inference speed of the two autoregressive counterparts, we propose a non-autoregressive variant (VPTR-NAR), which is inspired by the achitecture of DETR \cite{carion2020}. VPTR-NAR is illustrated in Fig. \ref{fig:NAR_model}. VPTR-NAR shares the same $\mathcal{T}_E$ as VPTR-PAR, while the $\mathcal{T}_D$ of them are slightly different. Firstly, target sequence for $\mathcal{T}_D$ is substituted with zero \cite{carion2020}, instead of $\{z_1, ... z_{L+N-1}\}$ generated by the CNN encoder, and a future frames query sequence $\{q_{L+1}, ..., q_{L+N}\}$ is fed into two different sublayers of $\mathcal{T}_D$, where $q_t\in \mathbb{R}^{H\times W\times C}, t\in [L+1, L+N]$. The future frame query sequence is randomly initialized and updated during training. Secondly, there is no temporal attention mask for any temporal attention layer of VPTR-NAR, because we do not need to impose conditional dependency among each future frame query. Theoretically, VPTR-NAR directly models the following conditional distribution:

\begin{equation}
    p(x_{L+N}, ..., x_{L+1}|x_{L}, ...,x_1)
\end{equation}

\textbf{Contrastive feature loss for VPTR-NAR.} Unfortunately, a combination loss of MSE and GDL, i.e., $\mathcal{L} = \sum_{t=L+1}^{L+N}\mathcal{L}_2(x_t,\hat{x_t}) + \mathcal{L}_{gdl}(x_t, \hat{x_t})$, is not enough to train VPTR-NAR. Specifically, we observe that all the predicted future frames of one video clip are somewhat similar to each other when VPR-NAR is trained with the same loss function as VPTR-FAR and VPTR-NAR, which indicates that VPTR-NAR cannot learn good motion information in this case. In NLP, a similar phenomenon is also observed for some non-autoregressive NMT models, where the Transformer decoder frequently generates repeated tokens \cite{wang2019non}. This is because autoregressive models make the estimation of joint distribution to be tractable and thus they are easier to train, even though slower. To deal with this problem, we maximize the mutual information between predicted future frame feature $\hat{z}_t$ and the future frame feature $z_t$ (ground-truth) generated by the CNN encoder by adapting another contrastive feature loss $\mathcal{L}_{c}$ \cite{andonian2021a}, where $t\in [L+1, L+N]$. $\mathcal{L}_{c}$ is defined by

\begin{equation}
    \mathcal{L}_{c}(z_t, \hat{z}_t) = \frac{1}{2}\sum_{s=1}^{S_l}l_{c}(\hat{v}_s, v_s, sg(\bar{v}_s)) + l_{c}(v_s, \hat{v}_s, sg(\hat{\bar{v}}_s)),
\end{equation}

\noindent where $v_s \in \mathbb{R}^{d_{model}}$ denotes a feature vector at spatial location $s$ of $z_t$, $\bar{v}_s \in \mathbb{R}^{(S_l - 1)\times d_{model}}$ denotes the collection of feature vectors at all other spatial locations of $z_t$. $\hat{v}_s$ and $\hat{\bar{v}}_s$ of $\hat{z}_t$ are defined in the same way. The total number of spatial locations in a feature map is $S_l = H\times W$, and $sg$ is the stop gradient operation. $l_c$ is the infoNCE-based contrastive loss formulated as follows,

\begin{multline}
    l_{c}(v, v^+, v^-) = \\
    -log\frac{exp(s(v, v^+))}{exp(s(v, v^+)) + \sum_{m=1}^Mexp(s(v, v^-))}
\label{eq: NCE}.
\end{multline}

Similar to other contrastive learning objectives, $(v, v^+)$ denotes a positive pair of examples and $(v, v^-)$ denotes a negative pair of examples. In detail, $v^+ \in \mathbb{R}^{d_{model}}$ is the spatially-corresponding ground-truth feature vector of a feature vector $v \in \mathbb{R}^{d_{model}}$, while $v^- \in \mathbb{R}^{M\times d_{model}}$ denotes the $M$ other ground-truth feature vectors at different spatial location. $s(v1, v2)$ is the feature dot-product similarity operation between feature vectors $v1$ and $v2$. We can finally compose the loss function of VPTR-NAR as

\begin{equation}
    \mathcal{L}_{NAR} = \sum_{t=L+1}^{L+N}\mathcal{L}_2(x_t,\hat{x_t}) + \mathcal{L}_{gdl}(x_t, \hat{x_t}) + \lambda_2 \mathcal{L}_{c}(z_t, \hat{z}_t).
\label{eq: NAR_loss}
\end{equation}

Different from VPTR-FAR and VPTR-PAR, VPTR-NAR predicts $N$ future frames simultaneously instead of recurrently. Besides, the testing behavior and training behavior of VPTR-NAR are the same. For the case that we need to predict more than $N$ future frames during test, we can use VPTR-NAR as a block-wise autoregressive model, i.e., taking the $N$ predicted future frames as past frames and feeding them back into the encoder for predicting the next $N$ future frames. 

\subsection{Training strategy} 
We divide the VFFP model training process into two stages. In stage one, the VPTR module is ignored and we only train the encoder and decoder as a normal autoencoder with the loss function of Eq. \ref{eq: AE_loss}, which aims to reconstruct all the frames of the whole training set perfectly. In stage two, we freeze the well-trained encoder and decoder and only optimize the parameters of the VPTR module. VPTR-FAR, VPTR-PAR and VPTR-NAR are trained with the loss functions defined in Eq. \ref{eq: FAR_loss}, Eq. \ref{eq: PAR_loss} and Eq. \ref{eq: NAR_loss} respectively. As we mentioned before, a two-stage training strategy is naturally compatible with the disentanglement of feature extraction and prediction process. Moreover, an end-to-end training of Transformers and the CNN autoencoder is much more difficult and it requires a much bigger memory for computations. Besides, we observe that a final joint finetuning of autoencoder and VPTR after two-stage training is not beneficial. Furthermore, the two-stage training strategy gives us a more flexible framework since we are allowed to test different VPTR variants without repetitive training of the encoder and decoder. 

\section{Experiments}
\subsection{Datasets and Metrics}
The proposed VPTR models are evaluated over three datasets: BAIR \cite{ebert2017}, KTH \cite{schuldt2004}, and MovingMNIST \cite{srivastava2015}. 

\textbf{BAIR} showcases video clips of a randomly moving robot arm that pushes different objects on a table. The training and testing sets of BAIR are predefined by the dataset authors. The frame size of BAIR is $64\times 64$. We normalize it for training, but without any data augmentation.

\textbf{KTH} includes grayscale video clips of different human actions. Following the experiments setup of previous work \cite{villegas2017a, jin2020}, the frames are center cropped to be square and then resize to $64\times 64$. Besides, frames without human are removed based on the results of a person detector. Video clips of persons 1-16 are used for training, and persons 17-25 are used for testing. The dataset pixels are normalized before training. Random horizontal and vertical flips of each video clip are utilized as data augmentation.

\textbf{MovingMNIST} is a synthetic dataset where two MNIST characters randomly move in a square. We utilize the same MovingMNIST dataset as E3D-LSTM \cite{wang2018a} instead of randomly generating frames on the fly. The frame size of MovingMNIST is $64\times 64$. The training set contains 10000 clips, test set contains 3000 clips and valid set contains 2000 clips. The same data augmentation method as KTH is applied to MovingMNIST dataset.

Following the experimental protocols of previous work, our VPTR models are trained to predict 10 future frames given 2 past frames for the BAIR dataset, and predict 10 future frames given 10 past frames for KTH and Moving MNIST.

\textbf{Metrics.} Learned Perceptual Image Patch Similarity (LPIPS) and Structural Similarity Index Measure (SSIM) are used to evaluate all the three datasets. Peak Signal-to-Noise Ratio (PSNR) is used to evaluate the KTH and BAIR dataset, and Mean Square Error (MSE) is used to evaluate the MovingMNIST dataset. All the LPIPS values are presented in $10^{-3}$ scale. Smaller values are better for LPIPS and MSE, while larger values are better for PSNR and SSIM.

\subsection{Implementation}
\textbf{Training stage one.} For KTH and BAIR, the output layer of the decoder is Tanh. For MovingMNIST, the output layer of decoder is Sigmoid. We set $\lambda_1 = 0.01$ for KTH and MovingMNIST, $\lambda_1 = 0$ for BAIR dataset. The optimizer is Adam, with betas of $(0.5, 0.999)$ and a learning rate of $2e^{-4}$. 

\textbf{Training stage two.} We define the visual features dimension of each frame as $H=8, W=8, d_{model}=528$. For local spatial MHSA, the local patch size is $K = 4$. VPTR-FAR includes 12 layers of VidHRFormer blocks. For VPTR-NAR and VPTR-PAR, the number of layers of $\mathcal{T}_E$ is 4, and the number of layers of $\mathcal{T}_D$ is 8. All Transformers are optimized by AdamW with a learning rate of $1e^{-4}$. Gradient clipping is employed to stabilize the training. $\lambda_2 = 0.1$ for the loss function of VPTR-NAR (Eq. \ref{eq: NAR_loss}).

\subsection{Results and discussion}

\textbf{Results on KTH.} We present the best results of all three VPTR variants in Table \ref{tab:kth-movingmnist}. The reported average metrics are calculated over 20 predicted frames, which follows the same evaluation protocol of previous works. Compared with some SOTA models, all three VPTR variants outperform them in terms of LPIPS by a large margin. For PSNR and SSIM, our proposed VPTR models reach competitive performances.

\begin{figure}[ht]
\centering
\includegraphics[clip, trim=5.7cm 3.8cm 4.7cm 3.5cm, width=\linewidth]{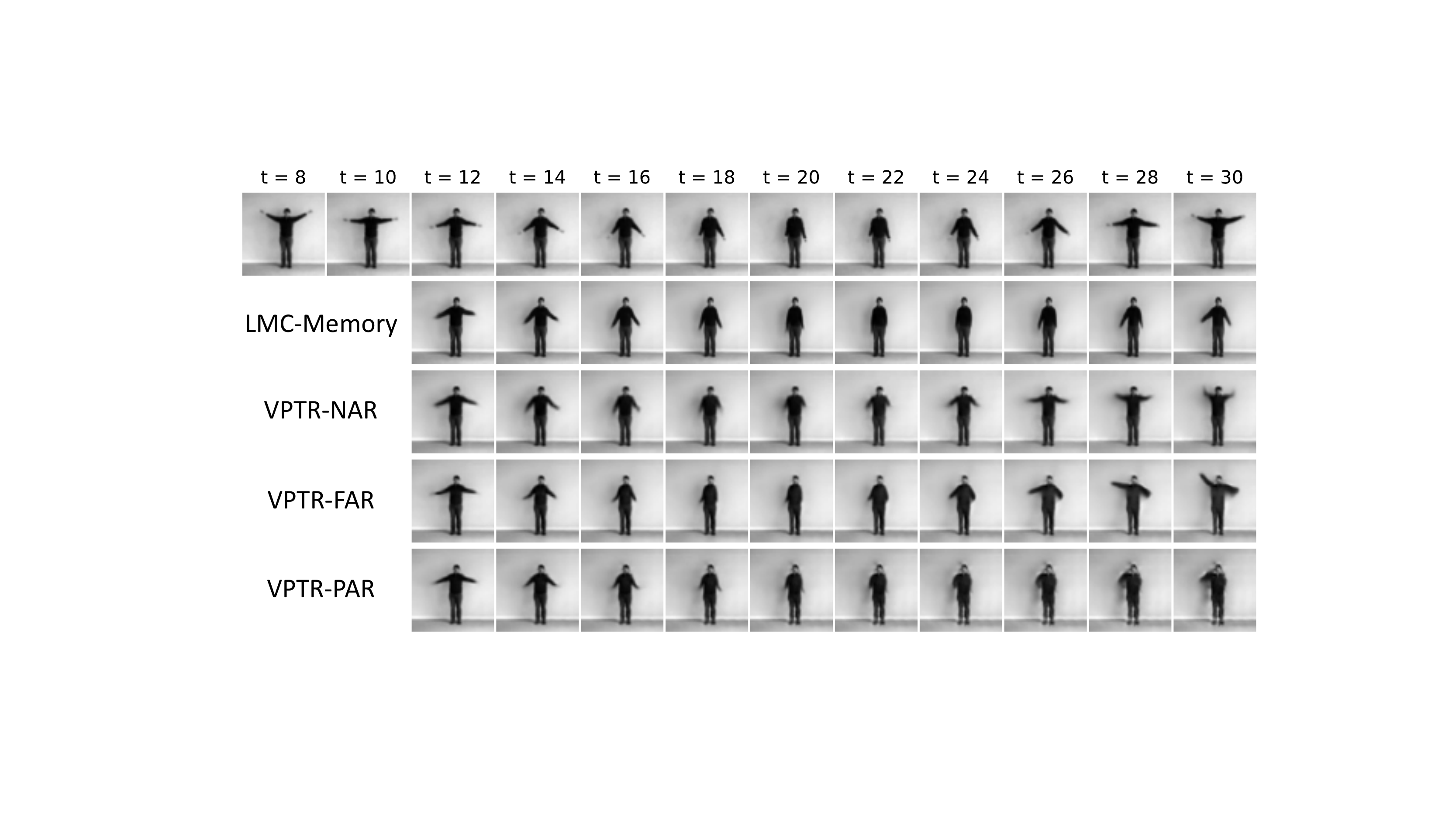}
\caption{Qualitative results on KTH dataset. The first row is ground-truth. For the past frames, $t\in [1, 10]$. For the future frames, $t\in [11, 30]$. }
\label{fig:KTH_results}
\end{figure}

Fig. \ref{fig:KTH_results} shows some prediction examples for a qualitative comparison. Comparing LMC-Memory with VPTR-NAR and VPTR-FAR, we observe that VPTR-NAR is better than VPTR-FAR and that both our VPTR models generate predictions that are more aligned with the ground-truth. It indicates that the VPTR-NAR and VPTR-FAR learns a better cyclic hand waving movements that is only condition on the past frames. We suspect that LMC-Memory retrieves some plausible but inaccurate motion from the learned memory bank and thus the prediction deviates from the ground-truth. Besides, due to a bigger accumulation of errors during inference, the last few predicted frames of both VPTR-PAR and VPTR-FAR are worse than VPTR-NAR. We  analyzed the reasons in detail at section \ref{ssec:ablation_study} and \ref{ssec: variants_compare}.

\begingroup
\setlength{\tabcolsep}{0.2pt}
\begin{table}[ht]
\caption{Results on KTH and MovingMNIST. $\uparrow$: higher is better, $\downarrow$: lower is better. \textbf{Boldface}: best results.}
\label{tab:kth-movingmnist}
\centering
\begin{tabular}{ccccccccc} \hline
\multicolumn{3}{c}{\multirow{3}{*}{Methods}} & \multicolumn{3}{c}{KTH} & \multicolumn{3}{c}{\begin{tabular}{@{}c@{}}Moving \\ MNIST \end{tabular}} \\
& & & \multicolumn{3}{c}{$10 \rightarrow20$} & \multicolumn{3}{c}{$10 \rightarrow 10$} \\
& & & PSNR$\uparrow$ & SSIM$\uparrow$ & LPIPS$\downarrow$ & MSE$\downarrow$ & SSIM$\uparrow$ & LPIPS$\downarrow$ \\ \hline
\multicolumn{3}{c}{MCNET \cite{villegas2017a}}& 25.95 & 0.804 & - & - & - & - \\
\multicolumn{3}{c}{PredRNN++ \cite{wang2018d}} & 28.47 & 0.865 & 228.9 & 46.5 & 0.898 & 59.5 \\
\multicolumn{3}{c}{E3D-LSTM \cite{wang2018a}} & 29.31 & 0.879 & - & \textbf{41.3} & 0.910 & - \\
\multicolumn{3}{c}{STMFANet \cite{jin2020}} & \textbf{29.85} & 0.893 & 118.1 & - & - & - \\
\multicolumn{3}{c}{Conv-TT-LSTM \cite{su2020a}} & 28.36 & \textbf{0.907} & 133.4 & 53.0 & 0.915 & \textbf{40.5} \\
\multicolumn{3}{c}{LMC-Memory \cite{lee2021}} & 28.61 & 0.894 & 133.3 & 41.5 & \textbf{0.924} & 46.9 \\ \hline
\multicolumn{3}{c}{VPTR-NAR} & 26.96 & 0.879 & 86.1 & 63.6 & 0.882 & 107.5 \\
\multicolumn{3}{c}{VPTR-PAR} & 25.40 & 0.836 & 84.8 & 93.2 & 0.859 & 138.4 \\
\multicolumn{3}{c}{VPTR-FAR} & 26.13 & 0.859 & \textbf{79.6} & 107.2 & 0.844 & 157.8 \\ \hline

\end{tabular}
\end{table}
\endgroup

\textbf{Results on MovingMNIST.} The results on the MovingMNIST dataset are shown in the right part of Table \ref{tab:kth-movingmnist}. In terms of SSIM, the performance of our VPTR models is close to the SOTA, but there are large gaps in terms of MSE and LPIPS, especially for VPTR-FAR and VPTR-PAR. After inspection of some prediction examples, we find that our VPTRs make poor predictions for the overlapping characters.

\begin{figure}[ht]
\centering
\includegraphics[clip, trim=5.5cm 10.5cm 4.7cm 3.5cm, width=\linewidth]{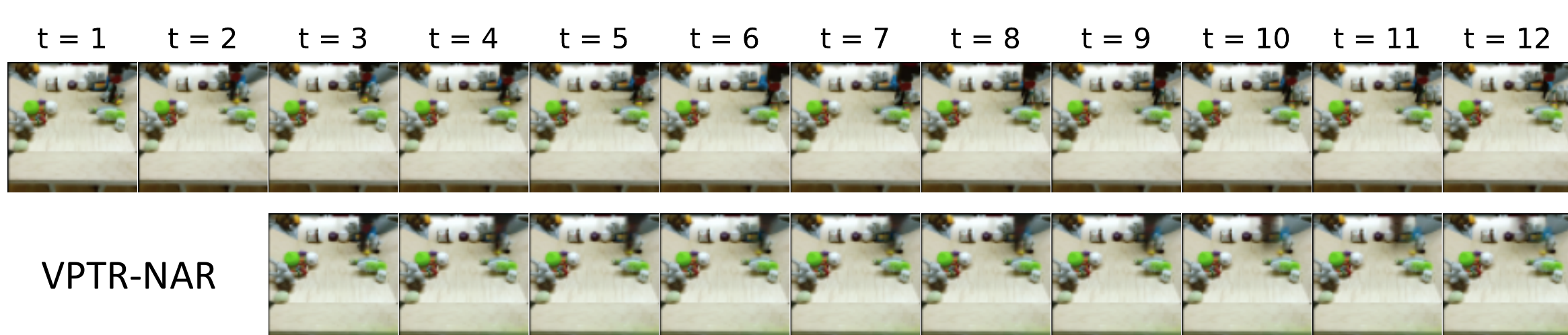}
\caption{Qualitative results on BAIR dataset. The first row is ground-truth. For the past frames, $t\in [1, 2]$. For the future frames, $t\in [3, 12]$. }
\label{fig:BAIR_results}
\end{figure}

\textbf{Results on BAIR.} Because the robot arm motion in BAIR dataset is random and we only condition on two past frames to prediction ten future frames, BAIR is a more challenging dataset than KTH and MovingMNIST. The test results on BAIR dataset are listed in Table \ref{tab:bair}. We find that the performance of VPTR-NAR is close to the reference models in terms of all three metrics. Particularly, VPTR-NAR outperforms two MCVD \cite{voleti2022} models in terms of both SSIM and PSNR, it also outperforms SVG-LP with regard to PSNR. We note that the predicted robot arm becomes blurry after the first few frames due to the deterministic nature of VPTRs, see Fig. \ref{fig:BAIR_results}. Our VPTRs could be extended to be stochastic models easily, and we expect that the stochastic version of VPTRs would achieve better performance on the BAIR dataset. The autoregressive variants have worse performances compared to the non-autoregressive variant because of the accumulation of errors, see the ablation study for more detailed analysis.

\begingroup
\setlength{\tabcolsep}{0.2pt}
\begin{table}[ht]
\caption{Results on BAIR. $\uparrow$: higher is better, $\downarrow$: lower is better. \textbf{Boldface}: best results.}
\centering
\begin{tabular}{cccccc} \hline
\multicolumn{3}{c}{\multirow{2}{*}{Methods}} & \multicolumn{3}{c}{$2 \rightarrow28$} \\
& & & PSNR$\uparrow$ & SSIM$\uparrow$ & LPIPS$\downarrow$ \\ \hline
\multicolumn{3}{c}{SV2P \cite{babaeizadeh2018}} & 20.36 & 0.817 & 91.4 \\
\multicolumn{3}{c}{MCVD-concat \cite{voleti2022}} & 17.70 & 0.797 & - \\
\multicolumn{3}{c}{MCVD-spatin \cite{voleti2022}} & 17.70 & 0.789 & - \\
\multicolumn{3}{c}{SVG-LP \cite{denton2018}} & 17.72 & 0.815 & 60.3 \\
\multicolumn{3}{c}{Improved VRNN \cite{castrejon2019}} & - & 0.822 & \textbf{55.0} \\
\multicolumn{3}{c}{STMFANet \cite{jin2020}} & \textbf{21.02} & \textbf{0.844} & 93.6 \\ \hline
\multicolumn{3}{c}{VPTR-NAR} & 17.77 & 0.813 & 70.0 \\
\multicolumn{3}{c}{VPTR-PAR} & 15.94 & 0.745 & 104.8  \\
\multicolumn{3}{c}{VPTR-FAR} & 15.76 & 0.724 & 110.7  \\ \hline
\end{tabular}
\label{tab:bair}
\end{table}
\endgroup

\subsection{Running time comparison}
\begin{figure}[ht]
\centering
\includegraphics[width=\linewidth]{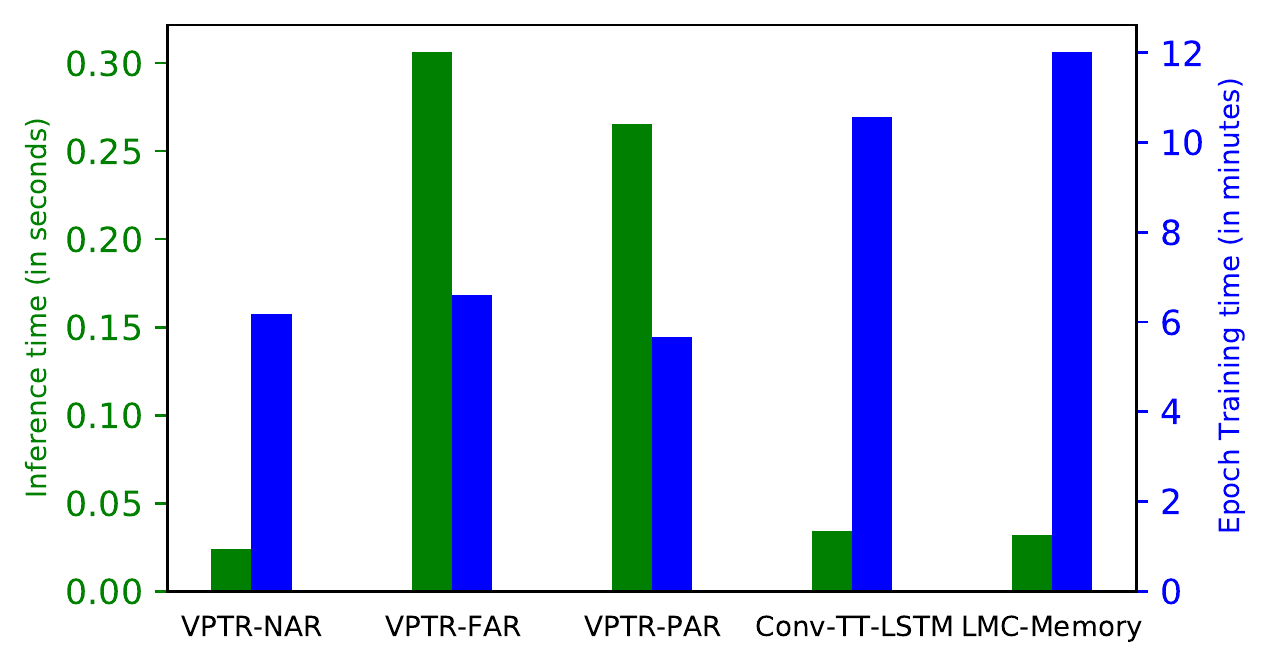}
\caption{Comparison of inference time and epoch training time.}
\label{fig:running_time}
\end{figure}

In order to show that VPTR is less complex and more efficient compared with ConvLSTM-based methods, we evaluated the inference time and epoch training time on the KTH dataset of all our VPTR variants and two SOTA ConvLSTM-based methods, Conv-TT-LSTM and LMC-Memory. Specifically, we measured the average inference time for predicting 10 future frames given 10 past frames. The results show that VPTR-NAR is the fastest one because of the non-autoregressive prediction. VPTR-FAR and VPTR-PAR are significantly slower than all other models because autoregressive prediction by Transformer is expensive. The results of average epoch training time indicate that all three VPTR variants have a similar training speed and are almost two times faster than the ConvLSTM-based models, thanks to thebenefit from the parallel computation of Transformers. On the contrary, ConvLSTM-based models need to slowly backpropagate through each time step during training. All models are tested on the same NVidia RTX3090 GPU with the same batch size.

\subsection{Ablation Study}
\label{ssec:ablation_study}

\begingroup
\setlength{\tabcolsep}{0.2pt}
\begin{table}[ht]
\caption{Ablation study on KTH and MovingMNIST. $\uparrow$: higher is better, $\downarrow$: lower is better. \textbf{Boldface}: best results.}
\centering
\begin{tabular}{ccccccccc} \hline
\multicolumn{3}{c}{\multirow{3}{*}{Methods}} & \multicolumn{3}{c}{KTH} & \multicolumn{3}{c}{\begin{tabular}{@{}c@{}}Moving \\ MNIST \end{tabular}} \\
& & & \multicolumn{3}{c}{$10 \rightarrow20$} & \multicolumn{3}{c}{$10 \rightarrow 10$} \\
& & & PSNR$\uparrow$ & SSIM$\uparrow$ & LPIPS$\downarrow$ & MSE$\downarrow$ & SSIM$\uparrow$ & LPIPS$\downarrow$ \\ \hline

\multicolumn{3}{c}{VPTR-NAR-BASE} & 26.92 & \textbf{0.881} & 94.6 & 64.2 & 0.880 & 114.2 \\
\multicolumn{3}{c}{VPTR-NAR-RPE} & \textbf{26.96} & 0.879 & 86.1 & \textbf{63.6} & \textbf{0.882} & \textbf{107.5} \\
\multicolumn{3}{c}{VPTR-NAR-FSTA} & 26.25 & 0.872 & 101.1 & 68.0 & 0.872 & 128.7 \\\hline
\multicolumn{3}{c}{VPTR-PAR-BASE} & 25.04 & 0.832 & 86.0 & 93.2 & 0.859 & 138.4 \\
\multicolumn{3}{c}{VPTR-PAR-RPE} & 25.40 & 0.836 & 84.8 & 104.2 & 0.848 & 139.4 \\
\multicolumn{3}{c}{VPTR-PAR-FSTA} & 24.33 & 0.814 & 89.9 & 81.6 & 0.873 & 105.9 \\
\multicolumn{3}{c}{VPTR-PAR-RIL} & 23.14 & 0.694 & 193.6 & 194.1 & 0.362 & 516.1 \\\hline
\multicolumn{3}{c}{VPTR-FAR-BASE} & 25.71 & 0.816 & \textbf{79.5} & 108.3 & 0.843 & 157.3 \\ 
\multicolumn{3}{c}{VPTR-FAR-RPE} & 26.13 & 0.859 & 79.6 & 107.2 & 0.844 & 157.8 \\ 
\multicolumn{3}{c}{VPTR-FAR-RIL} & 21.61 & 0.678 & 192.7 & 138.2 & 0.821 & 445.7 \\ \hline

\end{tabular}
\label{tab:ablation}
\end{table}
\endgroup

We conducted a thorough ablation study to investigate the influence of positional encodings, separated spatial and temporal attention and autoregressive inference methods. The ablation study results are summarized in Table \ref{tab:ablation}. Besides, we report the number of parameters and inference FLOPs of each different model in \ref{tab:ablation_flops}.

\textbf{RPE.} We take the VPTRs with fixed absolute positional encodings as the base models, i.e. VPTR-NAR-BASE, VPTR-PAR-BASE and VPTR-FAR-BASE. To investigate the influence of relative positional encodings for all VPTR variants, we substituted the 2D absolute positional encoding of all local spatial MHSA module with a learned 2D RPE, which give us VPTR-NAR-RPE, VPTR-PAR-RPE and VPTR-FAR-RPE. Our experiments show that in general, RPE is beneficial for the performance on both KTH and MovingMNIST datasets because the RPE models outperform the base models with regard to most metrics. Therefore, we can conclude that RPE is better than absolute positional encodings for VPTR. However, RPE would also introduce additional computational cost, see Table \ref{tab:ablation_flops}, which shows that FLOPs of RPE variants are slightly larger than the base models.

\textbf{Spatial-temporal separated attention.}
We propose to separate the spatial and temporal attention since we aim to reduce the complexity of the standard Transformer for video feature learning. However, this separated attention mechanism means that a feature at one location only attends to partial locations of the whole spatio-temporal space. In order to analyze the impact of the separated attention, the encoder-decoder attention layers of VPTR-NAR and VPTR-FAR are replaced with a full spatio-temporal attention (FSTA), which has a complexity of $\mathcal{O}(\frac{H^2W^2T^2}{P^2}d_{model})$. As we only replace the encoder-decoder attention layers, the increased computation cost is affordable. The FLOPs of FSTA variants shown in Table \ref{tab:ablation_flops} validate our arguments. We find that FSTA is not beneficial by comparing the VPTR-NAR-FSTA and VPTR-PAR-FSTA with their base counterparts. Consequently, it is safe to conclude that the alternate stacking of multiple spatial and temporal attention layers, as proposed with our VidHRFormer block, is capable of propagating global information from past frames to future frames.

\textbf{Autoregressive inference methods.} For two autoregressive variants, VPTR-FAR and VPTR-PAR, we can perform recurrently inference over latent space (RIL) or recurrently inference over pixel space (RIP) as we have mentioned in Section \ref{ssec: VPTR-FAR}. VPTR-FAR-BASE and VPTR-PAR-BASE are evaluated by RIP. The FLOPs presented in Table \ref{tab:ablation_flops} indicate that RIL variants require fewer FLOPs than the RIP models, i.e., it is a little faster than RIP. However, we observe that VPTR-FAR-RIL and VPTR-PAR-RIL are outperformed by a large margin due to the severe accumulation of errors of RIL.

Accumulation of errors exists for any inference process with autoregressive models that are trained with a teacher-force manner, and we analyze the detailed reasons in section \ref{ssec: variants_compare}. RIL gets a larger accumulated error than RIP because the VPTR-FAR and VPTR-PAR receive only supervision from the pixel space during training. According to the loss functions in Eq. \ref{eq: FAR_loss} and Eq. \ref{eq: PAR_loss}, it is clear that there is no direct penalty on the distance between the feature space generated by the CNN encoder and the feature space predicted by the Transformer. Furthermore, the pixel space dimensionality is smaller than the latent space dimensionality of the autoencoder, which is a common case for VFFP, as there are no skip connections from encoder to decoder and good reconstruction visual quality is expected. Therefore, we can hypothesize that recurrent inference solely by the Transformer predictor would make the predicted features quickly deviate from the ground-truth features (learned by the autoencoder during stage one). However, decoding the feature firstly and then encoding it back into latent space by the CNN encoder restricts the deviation to some degree.

\begingroup
\setlength{\tabcolsep}{2pt}
\begin{table}[ht]
\caption{Comparison of FLOPs and number of parameters for different models in the ablation study.}
\centering
\begin{tabular}{ccccc} \hline
\multicolumn{3}{c}{\multirow{3}{*}{Methods}} & \multicolumn{2}{c}{\begin{tabular}{@{}c@{}}Moving \\ MNIST \end{tabular}} \\
& & & \multicolumn{2}{c}{$10 \rightarrow 10$} \\
& & & FLOPs & $\#$Params \\ \hline

\multicolumn{3}{c}{VPTR-NAR-BASE} & 197.00G & 165.83M\\
\multicolumn{3}{c}{VPTR-NAR-RPE} & 201.13G & 162.48M \\
\multicolumn{3}{c}{VPTR-NAR-FSTA} & 197.81G & 165.83M  \\\hline
\multicolumn{3}{c}{VPTR-PAR-BASE} & 664.44G & 164.71M \\
\multicolumn{3}{c}{VPTR-PAR-RPE} & 679.59G & 164.87M \\
\multicolumn{3}{c}{VPTR-PAR-FSTA} & 668.93G & 164.71M \\
\multicolumn{3}{c}{VPTR-PAR-RIL} & 601.71G & 164.71M \\\hline
\multicolumn{3}{c}{VPTR-FAR-BASE} & 1.40T & 136.37M \\ 
\multicolumn{3}{c}{VPTR-FAR-RPE} & 1.45T & 133.02M \\ 
\multicolumn{3}{c}{VPTR-FAR-RIL} & 1.34T & 136.37M \\ \hline

\end{tabular}
\label{tab:ablation_flops}
\end{table}
\endgroup

\subsection{Comparison of VPTR variants}
\label{ssec: variants_compare}
\begin{figure}[ht]
\centering
\includegraphics[clip, trim=0.3cm 0.2cm 0.2cm 0.2cm, width=0.8\linewidth]{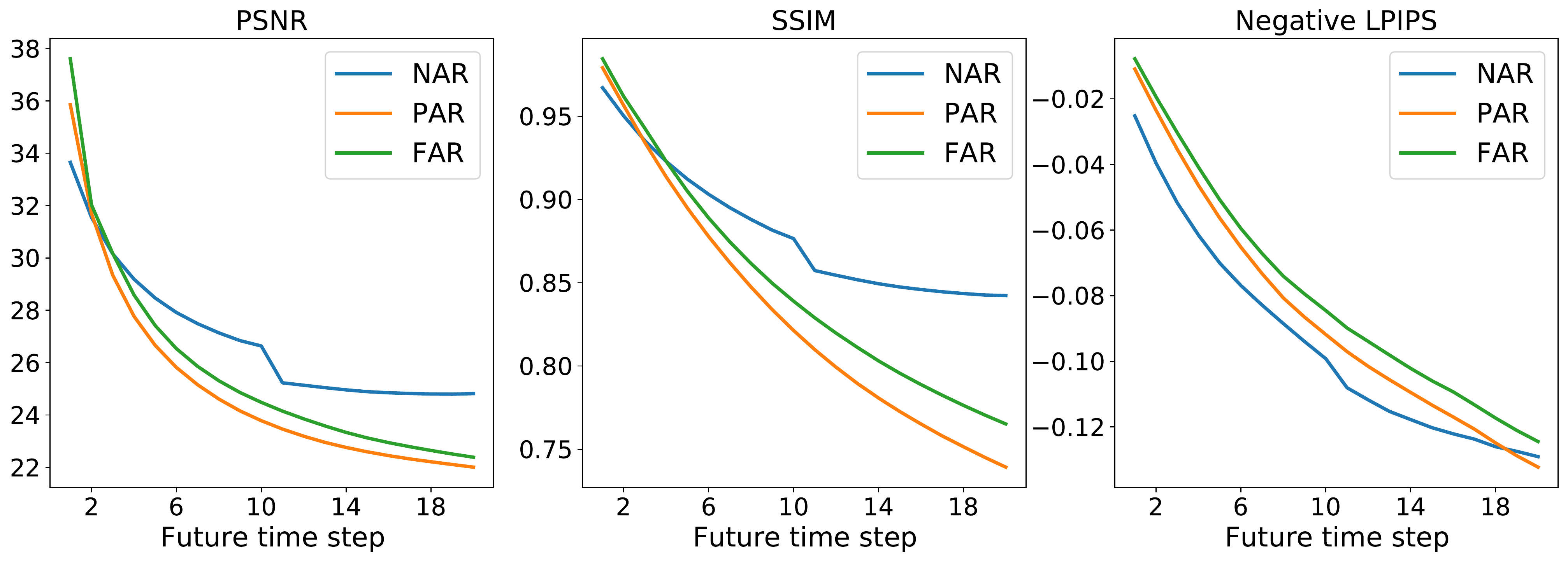}
\caption{Results of VPTR variants on KTH for increasing prediction steps.}
\label{fig:variants_comparison}
\end{figure}

For better visualization of the difference between the three VPTR variants, i.e., VPTR-NAR-BASE, VPTR-FAR-BASE and VPTR-PAR-BASE, the curves of the metrics with respect to the predicted future time steps are plotted in Fig. \ref{fig:variants_comparison}.  Recall that for LPIPS, a smaller value is better. The performance of VPTR-FAR and VPTR-PAR are close to each other, but VPTR-FAR is slightly better. Comparing the loss function of VPTR-FAR (Eq. \ref{eq: FAR_loss}) and VPTR-PAR (Eq. \ref{eq: PAR_loss}), the loss function of VPTR-FAR is calculated from $t=2$ to $t=L+N$, while VPTR-PAR is only trained by the loss that is calculated from $t=L+1$ to $t=L+N$. We believe the better performance of VPTR-FAR can be attributed to the larger supervision it receives during training. 

Comparing the autoregressive variants with VPTR-NAR, VPTR-FAR and VPTR-PAR achieve a better PSNR and SSIM than VPTR-NAR during the first few prediction steps, but their performance drops quickly due to the accumulation of errors introduced by the recurrent inference. The PSNR and SSIM curves demonstrate that VPTR-NAR has a smaller quality degradation. For the last 10 steps of the LPIPS curve, VPTR-NAR also has a smaller slope than VPTR-FAR. 

The accumulated inference errors of autoregressive VPTR variants are mainly due to the discrepancy between training and testing behaviors. Specifically, the previously predicted frames are used during inference instead of the ground-truth as during training, which leads to a worse generalization ability for VPTR-FAR or VPTR-PAR given the same number of VidHRFormer layers as VPTR-NAR. On the contrary, the training and testing behaviors of VPTR-NAR are the same. However, it is more difficult for the VPTR-NAR to directly estimate the joint distribution, so an additional contrastive feature loss and more parameters (the learnable future frame queries) are required.

VPTR-NAR has another advantage, i.e., a faster inference speed. Consider predicting $N$ future frames given $L$ past frames by the three VPTR variants. Then VPTR-NAR has a complexity of $\mathcal{O}(N^2)$, but the complexity for VPTR-FAR and VPTR-PAR is $\mathcal{O}(\sum_{n=1}^N n^2)$. Even though VPTR-PAR has the same complexity as VPTR-FAR in terms of the predicted future frames length, VPTR-PAR has a faster inference speed. Indeed, in VPTR-PAR, the past frames are processed by 4 layers of VidHRFormer block ($\mathcal{T}_E$), and each future frame are generated by passing through another 8 layers of VidHRFormer block ($\mathcal{T}_D$). However, in VPTR-FAR, each future frame is generated by passing all previous frames through 12 layers of VidHRFormer block, which costs more time. For simplicity, in this assessment, we ignored the $H, W$ and $d_{model}$ of video features, the computation cost of processing past frames, and we supposed that the predicted future frames length during test is the same as of the training. We tested the inference speed of three models on an NVidia RTX3090, the results in Figure \ref{fig:running_time} show that VPTR-NAR is $12.75$ times faster than VPTR-FAR, and VPTR-PAR is $1.2$ times faster than VPTR-FAR. The FLOPs results summarized in Table \ref{tab:ablation_flops} also validate our analysis.

\section{Conclusion}
In this paper, an efficient VidHRFormer block is proposed for spatio-temporal representation learning, and three different VFFP models are developed based on it. Our proposed VidHRFormer block could be applied to many other video processing tasks as a backbone. We compared the performance of the proposed VPTRs with SOTA models on various datasets, and our proposed methods are competitive with more complex ConvLSTM-based models. Finally, we conducted a through ablation study to analyze the influence of different modules for three VPTR variants, and we observed that VPTR-NAR achieves a better performance than VPTR-FAR and VPTR-PAR.









\bibliographystyle{unsrt}



\end{document}